\documentclass[11pt]{article} 
\usepackage{rldmsubmit,palatino}
\usepackage{graphicx}
\usepackage{amsmath}
\usepackage{algorithm}
\usepackage{algpseudocode}
\usepackage[capitalize]{cleveref}
\usepackage[font=small,labelfont=bf]{caption}

\DeclareMathOperator*{\argmax}{arg\,max}

\title{Decision Making in Non-Stationary Environments with Policy-Augmented Monte Carlo Tree Search}

\author{
Geoffrey Pettet \\
Vanderbilt University\\
Nashville, TN, 37212 \\
\texttt{geoffrey.a.pettet@vanderbilt.edu} \\
\And
Ayan Mukhopadhyay \\
Vanderbilt University \\
Nashville, TN, 37212 \\
\texttt{ayan.mukhopadhyay@vanderbilt.edu} \\
\And
Abhishek Dubey\\
Vanderbilt University\\
Nashville, TN, 37212 \\
\texttt{abhishek.dubey@vanderbilt.edu} \\
}

\begin{document}

\maketitle

\begin{abstract}

Decision-making under uncertainty (DMU), i.e., taking actions with uncertain outcomes using (potentially imperfect) observations, is present in many important problems. An open challenge is DMU in non-stationary environments, where the dynamics of the environment can change over time. Reinforcement learning (RL), a popular approach for DMU problems, learns a policy by interacting with a model of the environment offline. Unfortunately, if the environment changes the policy can become stale and take sub-optimal actions, and relearning the policy for the updated environment takes time and computational effort. An alternative is online planning approaches such as Monte Carlo Tree Search (MCTS), which perform their computation at decision time. Given the current environment, MCTS plans using high-fidelity models to determine promising action trajectories. These models can be updated as soon as environmental changes are detected to immediately incorporate them into decision making. However, MCTS’s convergence can be slow for domains with large state-action spaces. In this paper, we present a novel hybrid decision-making approach that combines the strengths of RL and planning while mitigating their weaknesses. Our approach, called \textit{Policy Augmented MCTS} (PA-MCTS), integrates a policy's action-value estimates into MCTS, using the estimates to "seed" the action trajectories favored by the search. We hypothesize that by guiding the search with a policy, PA-MCTS will converge more quickly than standard MCTS while making better decisions than the policy can make on its own when faced with nonstationary environments. We test our hypothesis by comparing PA-MCTS with pure MCTS and an RL agent applied to the classical CartPole environment. We find that PA-MCTS can achieve higher cumulative rewards than the policy in isolation under several environmental shifts while converging in significantly fewer iterations than pure MCTS.

\end{abstract}


\startmain 

\section{Introduction}\label{sec:intro}
\vspace{-0.15in}
Decision-making under uncertainty (DMU), i.e., taking actions with uncertain outcomes using (potentially imperfect) observations, is present in many important problems such as autonomous driving, emergency response, and medical diagnosis~\cite{kochenderfer2022algorithms}. An open challenge is DMU in non-stationary environments, where the dynamics of the environment can change over time. A decision agent must adapt to these changes or take sub-optimal actions. To illustrate, consider emergency response management (ERM), a domain we have significant experience with~\cite{mukhopadhyay2021review}. ERM deals with optimizing the allocation and dispatch of mobile resources such as ambulances in the face of uncertain incident demand to minimize response times. The state-action space for an ERM setting can be very large; for example, in a city with $20$ ambulances to allocate between $30$ potential waiting locations, there are $\text{\textit{Permutations}}(30,20) = \frac{30!}{10!} = 7.31\times10^{25}$ possible assignments at each decision epoch.  This complexity alone makes ERM a difficult problem, and non-stationarity compounds the challenge. The environments in which ERM systems operate shift over long time scales: road networks, congestion patterns, and demographics all change over time~\cite{mukhopadhyay2021review}. Sudden events such as road closures, inclement weather, and equipment failure can also unpredictably impact the environment. An effective ERM decision agent must adapt to such changes with minimal effects on service quality. 

There are two common DMU approaches: reinforcement learning (RL) and online planning. In RL approaches, an agent learns a policy $\pi$, a mapping from states to actions, through interacting with the environment. Often, learning takes place offline using environmental models, and once a policy is learned, it can be invoked nearly instantaneously at decision time. Deep RL methods have achieved state-of-the-art performance in many applications~\cite{silver_mastering_2016, kochenderfer2022algorithms}. However, when faced with non-stationary environments, a policy can become stale and make suboptimal decisions. Moreover, retraining the policy on the new environment takes time and considerable computational effort, particularly in problems with complex state-action spaces such as ERM. While RL methods specifically designed to operate in non-stationary environments have been explored~\cite{ortner2020variational, cheung2019non}, there is a delay between when a change is detected and when the RL agent converges to the updated policy. 

The alternative approach is to perform online planning using algorithms such as Monte Carlo Tree Search (MCTS). Given the current environment, these approaches perform their computation at decision time by planning using high-fidelity models to determine promising action trajectories. These models can be updated as soon as environmental changes are detected and then such changes can be immediately incorporated in decision-making. MCTS has been proven to converge to optimal actions given enough computation time~\cite{kocsis2006bandit}, but convergence can be slow for domains with large state-action spaces. The slow convergence is a particular issue for problem settings with tight constraints on the time allowed for decision-making, e.g., in ERM, when an incident occurs, any time used for decision-making increases the time until a responder is dispatched. In prior work, we have applied several approaches to scale MCTS to practical ERM problems, including decentralized~\cite{Pettet2020} and hierarchical planning~\cite{pettet2021hierarchical}, but these approaches make assumptions about the environment to prune the state-action space that can lead to sub-optimal decisions. 

Both RL and MCTS have weaknesses when applied to complex DMU problems in non-stationary environments. In this paper, we present a novel hybrid decision-making approach that combines the strengths of RL and planning while mitigating some of these weaknesses. The intuition behind our approach is that if the environment has not changed too much between when a policy was learned and when a decision needs to be made, the policy can still provide helpful information. Our approach, called \textit{Policy Augmented MCTS} (PA-MCTS), integrates a policy's action-value estimates into MCTS, using the estimates to "seed" the action trajectories favored by the search. The impact of the policy can be tuned based on how similar the environment at decision time is to the one used during training. We hypothesize that by guiding the search with a policy, PA-MCTS will converge more quickly than standard MCTS while making better decisions than the policy can make on its own. The primary advantage of this approach compared to pure RL approaches for non-stationary environments is that it does not require any relearning---it can be applied as soon as changes are detected. 

In this paper, we test our hypothesis by applying PA-MCTS to the OpenAI Gym Cartpole-v1 environment~\cite{brockman2016openai}. We first learn a policy using RL on the standard environment and then perform several experiments in which we vary parameters affecting environmental dynamics, such as the force of gravity and the cart's mass. Finally, we compare the performance of pure MCTS, the RL policy in isolation, and PA-MCTS. Our results show that PA-MCTS can achieve higher cumulative rewards than the policy in isolation under several environmental shifts while converging in significantly fewer iterations than pure MCTS. While this exploratory paper shows the promise of PA-MCTS, there are still open questions to consider: how can we determine the influence the policy should have at a decision time based on the current environment? Are there specific types of environmental shifts where this approach performs poorly? Are there bounds on how much the environment can change while still benefiting from PA-MCTS? In future work, we will explore these questions and apply the approach to more complex environments such as ERM.

\section{Approach}\label{sec:approach}
\vspace{-0.1in}
We consider a Markov Decision Process (MDP) which is represented by the tuple $(S,\mathcal{A},P(s,a),R(s,a))$ where $S$ is a finite state space, $\mathcal{A}$ is an action-space, $P(s,a)$ is a state transition function, and  $R(s,a)$ is a reward function defining the instantaneous reward for taking action $a$ in state $s$. We consider non-stationary environments, meaning that $P(s,a)$ and $R(s,a)$ can change over time. Our agent's goal is to choose the sequence of actions that maximizes the cumulative reward $G$ (referred to as the \textit{return}) received while interacting with the environment. We assume that a policy $\pi$ is learned offline under initial environmental conditions (defined by a transition and reward function). The policy can be used to compute the expected discounted reward of taking an action $a$ in a state $s$ and then following the policy (denoted by $Q(s,a)$).

Our approach is based on Monte Carlo Tree Search (MCTS), an iterative algorithm that builds a search tree online in an incremental and asymmetric manner. Each MCTS iteration consists of four stages: (1) selection, (2) expansion, (3) rollout, and (4) back-propagation. The stage relevant to our work is selection, in which the next node to expand is determined. Starting at the root node, a \textit{tree policy} is recursively applied to descend through the tree to pick the most promising child, continuing until it reaches a non-terminal leaf node to expand. A popular tree policy is the Upper Confidence Bounds for Trees (UCT) algorithm~\cite{kocsis2006bandit}, which selects the next node according to:
\begin{equation}
\argmax_{j \in \text{Children}(p)} \{ \overline{G}_{j} + c\sqrt{\text{ln}(n_p)/n_j}\}    
\end{equation}
where $p$ is the parent node, $\overline{G}_{j}$ is the average return of rollout simulations including child $j$, $n_p$ and $n_j$ are the number of times $p$ and $j$ have been visited respectively, and $c$ is a hyperparameter controlling the trade-off between exploitation and exploration.

Our contribution is a novel version of UCT that is modified to incorporate the policy's action-value estimates $Q(s,a)$ when estimating the value of a node, which we call \textit{Policy Augmented UCT} (PA-UCT). When PA-UCT is used as the tree policy for MCTS, we call the resulting algorithm \textit{Policy Augmented MCTS} (PA-MCTS). PA-UCT selects nodes to explore using the following policy
\begin{equation}
    \argmax_{j \in \text{Children}(p)} \ \ \ \alpha Q(s_p, a_{p,j}) + (1 - \alpha)\overline{G}_{j} + c\sqrt{\frac{\text{ln}(n_p)}{n_j}}
\end{equation}
where $s_p$ is the state at the parent node $p$, $a_{p, j}$ is the action which transitions state $s_p$ to child $j's$ state when taken, 
and $\alpha$ is a weight hyperparameter that controls the trade-off between the policy and rollout returns: if $\alpha = 1$, PA-UCT reduces to greedy action selection based on the learned policy, whereas if $\alpha = 0$, it reduces to standard UCT. If $\alpha \in (0, 1)$, then both estimates are considered for action selection. 

MCTS is proven to converge to the optimal action given infinite iterations~\cite{kocsis2006bandit}. As the tree search produces improved estimates with increasing iterations given the updated environmental dynamics, we can decrease the influence of the policy on PA-UCT. Therefore, an alternative version of PA-UCT, PA-UCT with $\alpha$-decay, uses the following selection policy on iteration $i$:
\begin{equation}
    \argmax_{j \in \text{Children}(p)} \ \ \ \alpha(\frac{1}{1+k \cdot i})Q(s_p, a_{p,j}) + (1 - \alpha(\frac{1}{1+k \cdot i}))\overline{G}_{j} + c\sqrt{\frac{\text{ln}(n_p)}{n_j}}
\end{equation}
where $k$ is a hyperparameter that controls the rate of decay. 
\section{Experiments}\label{sec:exp}
\vspace{-0.1in}
To evaluate the efficacy of PA-MCTS, we first learn a policy $\pi$ for the OpenAI Gym cartpole-v1 environment~\cite{brockman2016openai}, using a double deep Q RL algorithm~\cite{van2016deep}. Environment parameters are set to default values, except the maximum length of each episode which is increased from 500 to 2500 time steps, in order to evaluate the performance of PA-MCTS\footnote{The relevant default parameters are that gravity $g = 9.8$ and the cart's mass $m = 1.0$. The standard cartpole reward function of returning $R(s,a) = 1.0$ for each time step before reaching a terminal state was used. The double Q-learning agent's learning rate $= 0.001$, and a Boltzmann control policy was used. The agent learned for 300000 steps}.

\begin{figure}
    \centering
    \includegraphics[width=0.9\columnwidth]{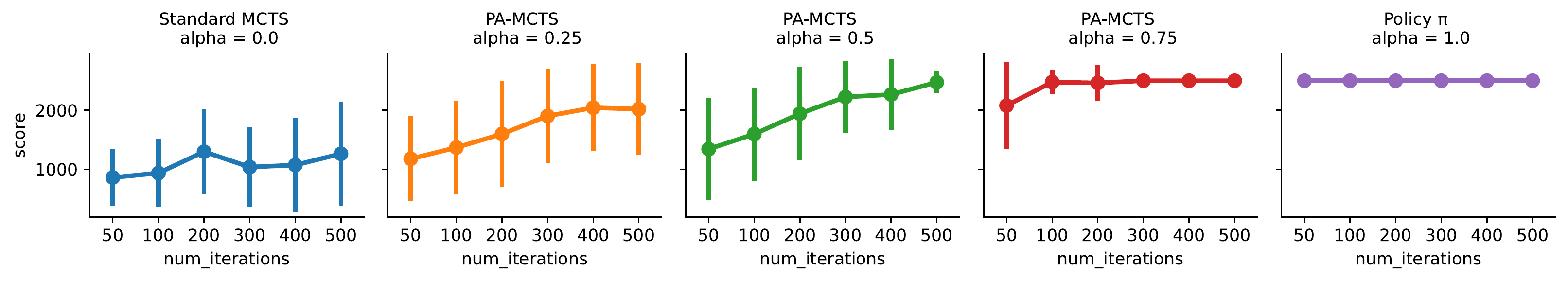}
    \caption{PA-MCTS results when applied to the default CartPole environment. Columns represent values for $\alpha$. The individual plots' x-axis is the number of MCTS iterations per decision epoch. The y-axis is the cumulative reward. The dots represent the mean cumulative reward over 50 samples, while the vertical lines are the standard deviation.}
    \label{fig:default}
\end{figure}

We then perform experiments comparing $\pi$ ($\alpha=1.0$) and MCTS ($\alpha=0.0$) \footnote{Recall that when $\alpha = 1.0$, PA-UCT reduces to greedy action selection based on the learned policy, and when $\alpha = 0.0$, it reduces to standard UCT.} against PA-MCTS\footnote{We use the following PA-MCTS hyperparameters in all experiments: the exploration-exploitation tradeoff parameter $c = 50$, the planning horizon is 500 time steps, the discount factor $\gamma = 0.999$, and the decay rate $k = 0.0$} with several values for $\alpha$ in the default environment used to learn $\pi$. We use the following MCTS iteration budgets to evaluate the convergence of each approach: $\{$50, 100, 200, 300, 400, 500$\}$. The results are shown in figure 1. Our first observation is that the policy $\pi$ ($\alpha=1.0$) achieves the best possible return of 2500, as expected. We also observe that PA-MCTS with $\alpha \in \{0.25, 0.5, 0.75\}$ converges in far fewer iterations than standard MCTS (i.e. with $\alpha = 0.0$), with $\alpha = 0.75$ converging to the optimal return at 300 iterations.

\begin{figure}
    \centering
    \includegraphics[width=0.9\columnwidth]{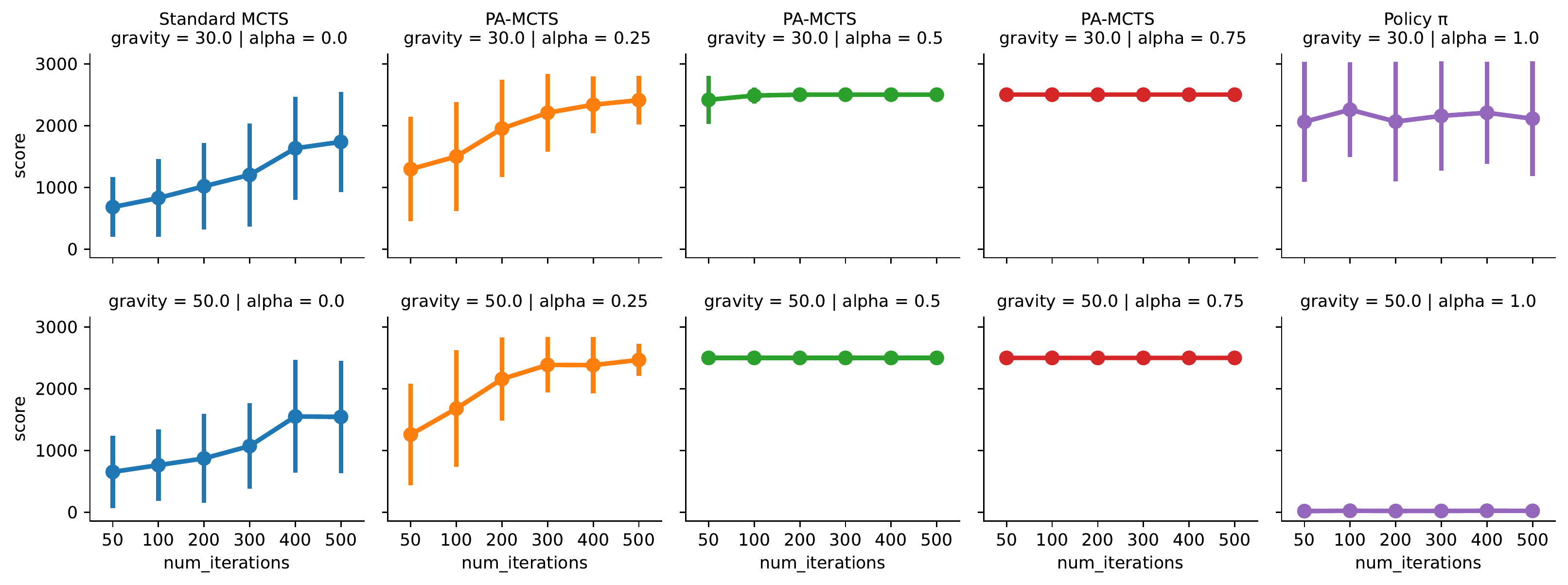}
    \caption{PA-MCTS results with gravity modified from its default value of 9.8 m/s$^2$. Rows represent different gravity values. Columns represent values for $\alpha$. The individual plots' x-axis is the number of MCTS iterations per decision epoch. The y-axis is the cumulative reward. The dots represent the mean cumulative reward over 50 samples, while the vertical lines are the standard deviation.}
    \label{fig:gravity}
\end{figure}

\begin{figure}
    \centering
    \includegraphics[width=0.9\columnwidth]{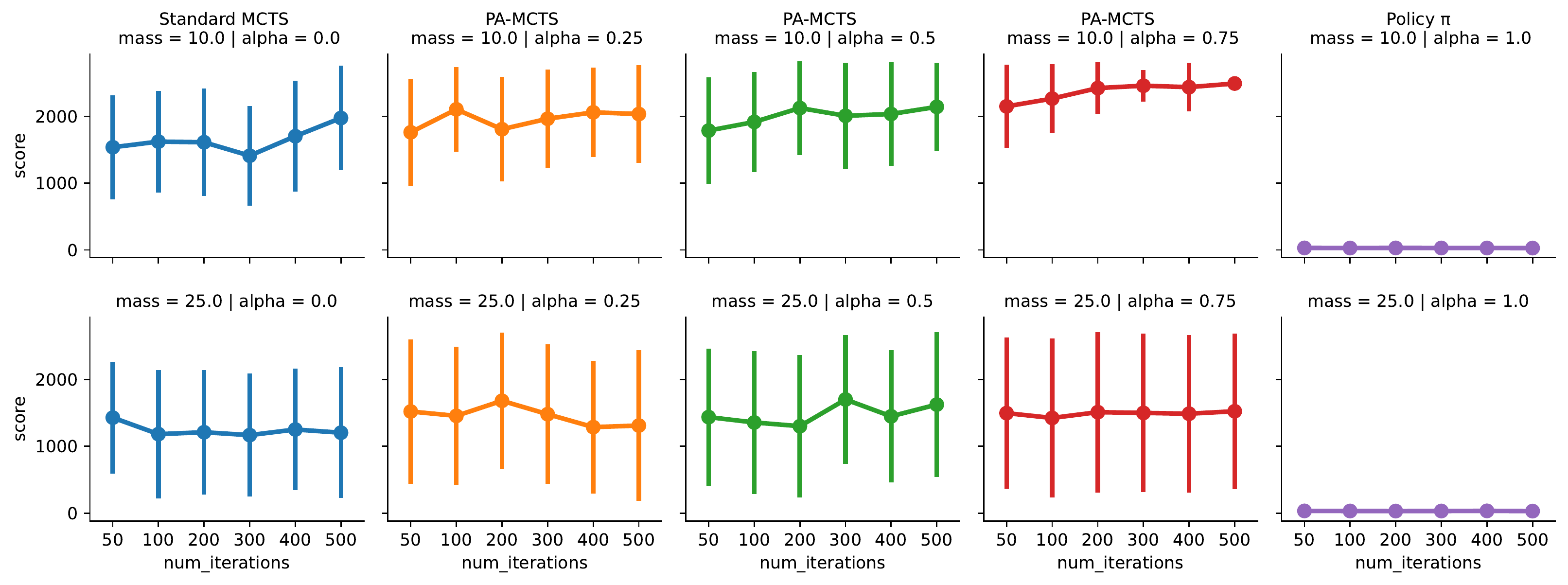}
    \caption{PA-MCTS results with the cart's mass modified from its default value of 1.0 kg. Rows represent different cart masses. Columns represent values for $\alpha$. The individual plots' x-axis is the number of MCTS iterations per decision epoch. The y-axis is the cumulative reward. The dots represent the mean cumulative reward over 50 samples, while the vertical lines are the standard deviation.}
    \label{fig:mass}
\end{figure}

\begin{figure}
    \centering
    \includegraphics[width=0.9\columnwidth]{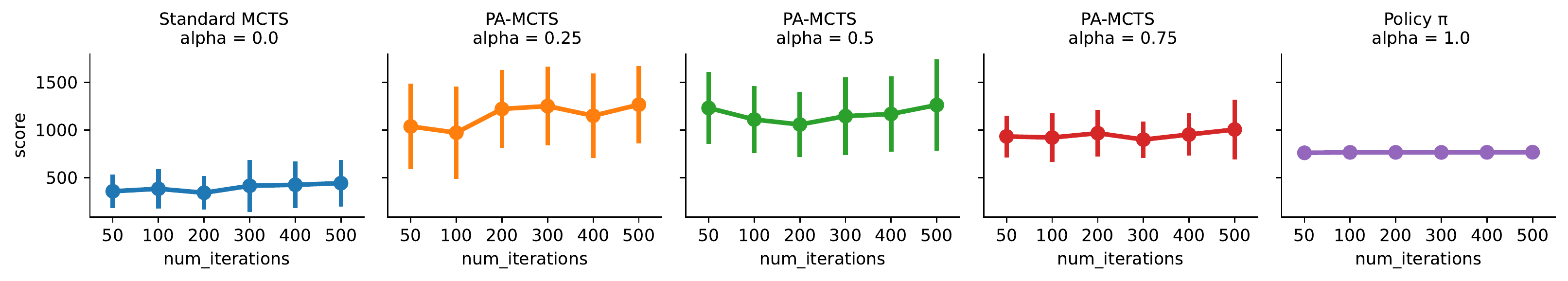}
    \caption{PA-MCTS results with a modified reward function which incentivizes staying near the center of the track. Columns represent values for $\alpha$. The individual plots' x-axis is the number of MCTS iterations per decision epoch. The y-axis is the cumulative reward. The dots represent the mean cumulative reward over 50 samples, while the vertical lines are the standard deviation.}
    \label{fig:mu_xp}
\end{figure}

The second set of experiments introduce non-stationarity to the environment by modifying two environmental parameters: we change the gravitational constant $g$ from the default value of 9.8 m/s$^2$ to 30.0 m/s$^2$ and 50.0 m/s$^2$, and change the cart's mass $m$ from the default of 1.0 kg to 10.0 kg and 25.0 kg. The results are shown in figures~\ref{fig:gravity} and \ref{fig:mass}. Our first observation is that the policy's performance in isolation degrades significantly: with $g =$ 30 m/s$^2$ and $\alpha = 1.0$, the mean return does not reach the optimal value, while for $g =$ 50.0 m/s$^2$ and $m \in \{$10.0 kg, 25.0 kg$\}$ the policy obtains returns close to 0. We also observe that PA-MCTS again converges in significantly fewer iterations than standard MCTS in most cases, notably achieving the optimal return within 50 iterations for $g \in \{$30.0 m/s$^2$, 50.0 m/s$^2\}$ and $\alpha = 0.75$ in figure \ref{fig:gravity}. The exception is $m =$ 25.0 kg in figure \ref{fig:mass}, where PA-MCTS does not significantly improve upon standard MCTS with any tested $\alpha$. There are several possible explanations for this: we may not have tested the optimal $\alpha$ value, the environment may have shifted too much for the policy to be useful, or it may be that balancing the pole with a cart this massive is too challenging given the default action-space. More study is needed to determine the amount of environmental shift that PA-MCTS can handle without retraining the policy. Finally, we evaluate PA-MCTS with a modified reward function. In the standard cartpole environment, the agent receives a reward of 1 for each time-step before the episode is terminated. We modify the reward function to return $1- (|x_p|/X_t)$, where $x_p$ is the cart's $x$ coordinate and $X_t$ is the track's boundary. In the cartpole environment, the center of the track is at $x = 0$ and the boundaries are at $x = -X_t$ and $x = X_t$. Therefore, this new reward function provides higher rewards the closer the cart is to the center of the track. The results with this reward function are shown in figure \ref{fig:mu_xp}. We again observe that PA-MCTS obtains higher returns than the policy while converging in fewer iterations than standard MCTS. 

\section{Conclusion}\label{sec:conclusion}
\vspace{-0.1in}
We present a novel hybrid decision-making approach that combines the strengths of reinforcement learning and planning for non-stationary environments called \textit{Policy Augmented MCTS} (PA-MCTS). Using the classical cartpole environment, we show that PA-MCTS can achieve higher cumulative rewards than an RL agent in isolation under environmental shifts while converging in significantly fewer iterations than pure MCTS. In future work, we will apply PA-MCTS to more complex problems, further explore the effect of hyperparameters such as the policy's influence weight $\alpha$ and the decay rate $k$, and study how we can use observations of the current environment to determine $\alpha$ at decision time. 

\bibliographystyle{plain} 
\bibliography{refs}

\end{document}